# Effect of roundabout design on the behavior of road users: A case study of roundabouts with application of Unsupervised Machine Learning

Tasnim M. Dwekat[1], Ayda A. Almsre[2]

[1] Al-najah University,

Nablus, Palestine.

(E-mail: s12028850@stu.najah.edu)

[2]Al-najah University,

Nablus, Palestine.

(E-mail: s12029716@stu.najah.edu)

Course number: 10672224

Course name: Artificial Intelligence

Dr. Huthaifa I. Ashqar

24 May 2022

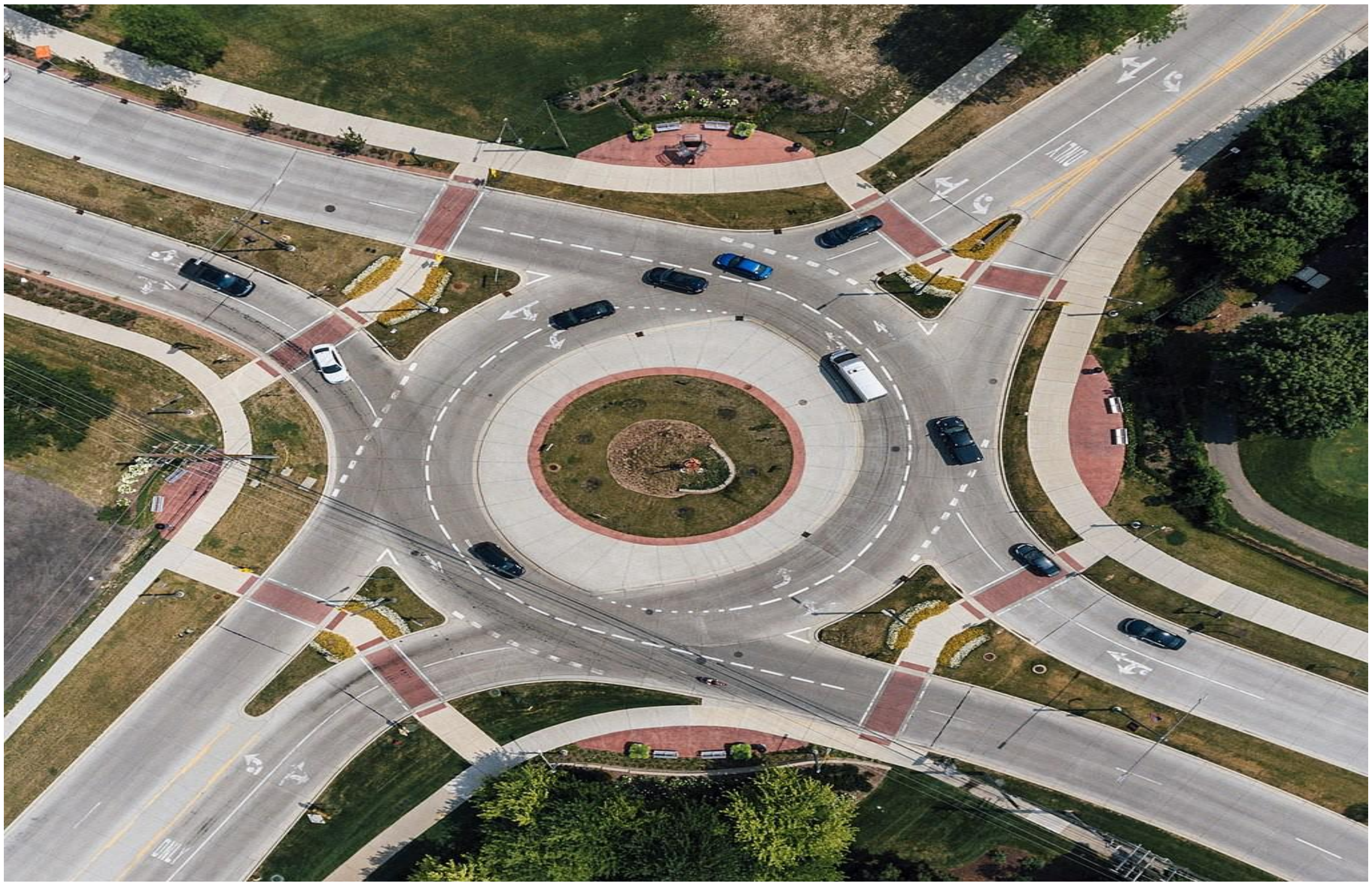

Fig. 1:An exemplary visualization of a given traffic scene included roundabouts in the rounD dataset. Traffic recordings taken by a drone

## ABSTRACT

This research aims to evaluate the performance of the rotors and study the behavior of the human driver in interacting with the rotors.
In recent years, rotors have been increasingly used between countries due to their safety, capacity, and environmental advantages, and because they provide safe and fluid flows of vehicles for transit and integration. It turns out that roundabouts can significantly reduce speed at twisting intersections, entry speed and the resulting effect on speed depends on the rating of road users. In our research, (bus, car, truck) drivers were given special attention and their behavior was categorized into (conservative, normal, aggressive). Anticipating and recognizing driver behavior is an important challenge. Therefore, the aim of this research is to study the effect of roundabouts on these classifiers and to develop a method for predicting the behavior of road users at roundabout intersections.

Safety is primarily due to two inherent features of the rotor. First, by comparing the data collected and processed in order to classify and evaluate drivers' behavior, and comparing the speeds of the drivers (bus, car and truck), the speed of motorists at crossing the roundabout was more fit than that of buses and trucks. We looked because the car is smaller and all parts of the rotor are visible to it. So drivers coming from all directions have to slow down, giving them more time to react and mitigating the consequences in the event of an accident.
Second, with fewer conflicting flows (and points of conflict), drivers only need to look to their left (in right-hand traffic) for other vehicles, making their job of crossing the roundabout easier as there is less need to split attention between different directions.

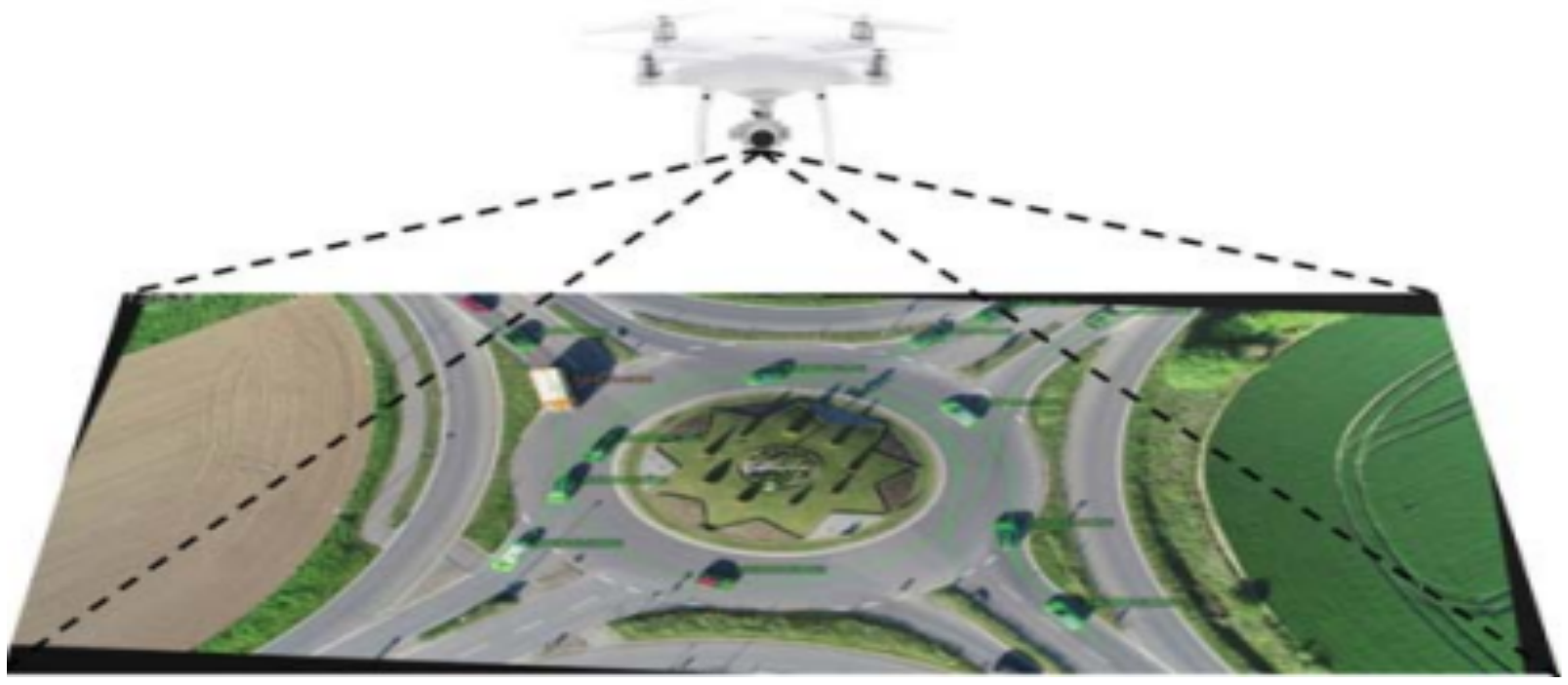

Fig. 2: We propose to use a camera-equipped drone to record traffic at roundabouts. The trajectory and

The class of each road user is extracted using computer vision algorithms.

# INTRODUCTION

In our research, we talk about autonomous driving as one of the main applications of many areas of research. In recent years there has been a growing interest in understanding the differences between road users (such as drivers, pedestrians, and cyclists), which is what we focus on in this paper, with references to three patterns: Using **aggressive**, **normal** and **conservative** features extracted from vehicle kinematics.

This interest helps develop new technologies for autonomous driving: it can give insight into how humans behave in each scenario, how to learn from human reactions, how to form an autonomous, human team and it can also provide a more realistic simulation environment for autonomous car developers.

To make the terms more understandable, here are some definitions of the most frequently used terms In the research,:

A self-driving vehicle, also known as an autonomous vehicle (AV), self-driving vehicle, driverless vehicle, or automated vehicle (automated vehicle), [1] [2] [3] is a vehicle comprising vehicle automation, which is, A ground vehicle capable of sensing its environment and moving safely with little or no human input. [4] [5].

Conservative drivers tend to adopt relatively conservative and conservative driving strategies, the vehicle's movement characteristics will be relatively stable, that is, there is no danger, if the speed limit is 70 and will reach 65, there is no chance of traction because of the speed, this type of driver has a higher commitment than others At traffic lights, traffic lights, etc.

As for the normal driver, who is the one-speed driver, who often travels at a constant average speed and does not change it often, he adheres to most of the signals, often bypassing the yellow ones.

Aggressive driving is defined by the National Highway Traffic Safety Administration as the behavior of an individual who "commits a range of moving traffic offenses to endanger other persons or property", driving at high speeds and bypassing most traffic lights. [6]

Nowadays, the ideas of smart cities and autonomous/automated driving are closely related to linking vehicles with other participants in traffic, smart infrastructure and cloud computing platforms through so-called electronic physical frameworks (CPFs) [7].

The main challenge of the ego vehicle is to anticipate the traffic scene surrounding the ego vehicle. Anticipating the surrounding traffic landscape is particularly difficult on roundabouts due to the interdependence of the respective road users.

One way to address this challenge is to use increasing amounts of data. There are many data sets available to support these studies [8, 9, 10, and 11]. Which is causing an increase in the popularity of data-driven methods that rely on large-scale datasets in recent years [12].

We obtained the data for road users, and ultimately chose car, truck and bus drivers, because the focus of this study is on the effect of rotors on these categories. In the future, we aspire to develop self-driving cars to become easier to deal with roundabouts, our simple observation was from this data, that cars are the most fit and quick to cross and deal with the roundabout, as for trucks, they are the most difficult and most time consuming in crossing, and the bus, the average among them.

We also noticed that the vehicle's speed gradually decreases when they reach the roundabouts. The rotors significantly reduce the speed at the intersections and on the links between the rotors, and here lies the main benefit of the rotor, as this contributes to the reduction of the number of accidents. Conflict studies indicated an overall reduction in accident risk of 44%. [13] Risks to vulnerable road users were significantly reduced, while there was no reduction for vehicle occupants. According to the Spanish Traffic Department (DGT) [14], “Circular lanes are a special type of intersection where they are connected by a ring that defines the flow of periodic traffic around a central island.” Because of the complex maneuvers [15, 16] of self-driving vehicles. Moreover, this type of traffic infrastructure provides a lot of information that can be extracted from images, such as paths or locations of vehicles.

For the data, the required data cannot be generated by methods in simulation or by systematic tests on the test path, because the recorded behavior will be biased and will not include corner states likely to occur in real-world conditions. However, if the data collected on public roads includes typical misconduct such as speeding or reckless driving, the data must be collected in an efficient manner to provide a large enough quantity to develop and validate automated driving systems [17].

We've got data ready, we just processed it and put some rules and compilation on it.

In our research, we used the dataset as a basis for road user, lane prediction and driver models. We determine driving behavior on different roads with different lighting conditions [18] and plan a safe and comfortable route for all road users.

A study has been proposed in which traffic is monitored at a roundabout intersection, with information about vehicle speeds, steering wheel angle and cornering geometry obtained from vehicle routes. Furthermore, the information includes the number of dynamic objects, the presence of pedestrians, cyclists and motorcyclists, and the distances between vehicles.

A solution that combines artificial intelligence and image processing technologies using a drone equipped with a camera to control when vehicles enter the roundabout [19].

Knowing the effect of the roundabouts on road users, these results can be used to develop the design of the roundabouts, in choosing their locations and sizes, and improving the design of the roundabout to suit the types of driving that drivers follow to reduce road accidents. Courses can be allocated to specific road users, and ego cars can also be developed to work with a better algorithm on roundabouts or allocate their own paths. These results can also be used in research to save fuel or fuel consumption rates for vehicles, and to enhance the movement of pedestrians or bicyclists on the roads, after knowing their impact. On cars, buses and trucks.

## RELATED WORK

In recent years, many data sets have been published to solve tasks in the field of automated driving and to study the importance and impact of roundabouts on road users. They can be roughly categorized into cognition and trajectory datasets.

The Five Roundabouts dataset [20] is a dataset published in 2019. Using a total of six Ibeo lidar vehicle scanners parked near roundabouts, more than 23,000 vehicles were tracked at five undisclosed intersections in Australia. Since the focus of their work is on analyzing vehicle behavior at roundabouts, no pedestrians were tracked for this data set, which is close to our research since we don't care about pedestrians, at this point.

Because the tracking system had a limited detection range and perspective-related obstruction, vehicles entering from side roads are not considered.

As the need for data sets grows due to the growing popularity of data-driven approaches to automated driving, the number of methods needed to create them is also growing. The development began with the creation of datasets sensitive to perception problems, for example: all dynamic road users are annotated by bounding boxes or the static infrastructure is intrinsically partitioned.

Several studies have been conducted to apply drone technology in traffic monitoring and different types of drones are used or tested to measure traffic data. Toth and others. [21] Examines the quality of traffic data obtained through the operation of an unmanned helicopter, and aerial photographs. Bury et al. It collects real-time data for traffic monitoring, pattern assessment and analysis, which leads to statistical errors in measurements due to drone instability and limited data collection only when the drone is in motion [22]. One potential of drones is to enhance traffic control. Traditional methods and traffic monitoring systems are only used to monitor simple traffic statistics at specific locations, while rural comprehensive traffic is excluded, mainly due to cost-effectiveness. On the other hand, drones appear to be a cost-effective solution that covers traffic control requirements in rural areas.

In addition, a lot of research has been done to analyze circular traffic flow using data obtained from various sources. St Aubin et al1 used traffic data from a fixed video camera system to analyze driver behavior at roundabouts in Canada. We extracted various traffic parameters by interpreting vehicle paths. Similarly, Mussone et al. attempted to analyze rotation performance by applying image processing techniques to still video camera data. All of these studies have devised and demonstrated different ways of evaluating the performance of rotors.

However, the main objective in our research is to assess drivers' behavior and analyze the risks and effects for vertigo.

Errors in vehicle detection and tracking by drones may occur for various reasons such as partial obstruction, shadows, nearby objects, false detections, or, as we mentioned previously, poor shooting angle or lack of clarity. Studies of how to track objects in videos from drones also believe that errors that occur during processing are due to the inability to detect objects that vary in size significantly during their passage or that move very quickly [23]. Therefore, processing software algorithms must be robust in the face of these many situations [24].

```
In [5]: tracks
```

Out[5]:

| | recordingId | trackId | frame | trackLifetime | xCenter | yCenter | heading | width | length | xVelocity | yVelocity | xAcceleration | yAcceleration | lonVelocity |
|---|---|---|---|---|---|---|---|---|---|---|---|---|---|---|
| **0** | 0 | 0 | 25218 | 0 | 141.0003 | -138.8970 | 107.7517 | 0.8158 | 1.0443 | -0.2620 | 0.7691 | -0.0941 | 0.5095 | 0.8123 |
| **1** | 0 | 0 | 25219 | 1 | 140.9896 | -138.8657 | 107.7339 | 0.8158 | 1.0443 | -0.2655 | 0.7899 | -0.0963 | 0.5187 | 0.8332 |
| **2** | 0 | 0 | 25220 | 2 | 140.9791 | -138.8335 | 107.7124 | 0.8158 | 1.0443 | -0.2691 | 0.8110 | -0.0987 | 0.5276 | 0.8544 |
| **3** | 0 | 0 | 25221 | 3 | 140.9686 | -138.8011 | 107.6877 | 0.8158 | 1.0443 | -0.2730 | 0.8326 | -0.1013 | 0.5361 | 0.8762 |
| **4** | 0 | 0 | 25222 | 4 | 140.9579 | -138.7678 | 107.6599 | 0.8158 | 1.0443 | -0.2770 | 0.8548 | -0.1038 | 0.5441 | 0.8986 |
| **...** | ... | ... | ... | ... | ... | ... | ... | ... | ... | ... | ... | ... | ... | ... |
| **116033** | 1 | 268 | 10897 | 357 | 169.2958 | -25.5921 | 52.6207 | 1.9505 | 4.7414 | 7.4750 | 9.7503 | 1.0161 | 0.7268 | 12.2859 |
| **116034** | 1 | 268 | 10898 | 358 | 169.5951 | -25.2026 | 52.5659 | 1.9505 | 4.7414 | 7.5149 | 9.7792 | 1.0156 | 0.7270 | 12.3331 |
| **116035** | 1 | 268 | 10899 | 359 | 169.8953 | -24.8118 | 52.5087 | 1.9505 | 4.7414 | 7.5553 | 9.8084 | 1.0155 | 0.7273 | 12.3809 |
| **116036** | 1 | 268 | 10900 | 360 | 170.1977 | -24.4188 | 52.4666 | 1.9505 | 4.7414 | 7.5960 | 9.8377 | 1.0155 | 0.7273 | 12.4289 |
| **116037** | 1 | 268 | 10901 | 361 | 170.5024 | -24.0243 | 52.4258 | 1.9505 | 4.7414 | 7.6366 | 9.8668 | 1.0155 | 0.7273 | 12.4768 |

2964145 rows × 17 columns

**Table one: Overview of datasets of recorded paths used at circular intersections.**

| | index | recordingId | trackId | initialFrame | finalFrame | numFrames | width | length | class |
|---|---|---|---|---|---|---|---|---|---|
| **0** | 0 | 0 | 0 | 25218 | 25715 | 498 | 0.8158 | 1.0443 | pedestrian |
| **1** | 1 | 0 | 1 | 25118 | 25715 | 598 | 0.5949 | 0.8228 | pedestrian |
| **2** | 2 | 0 | 2 | 25043 | 25689 | 647 | 0.6776 | 1.2047 | pedestrian |
| **3** | 3 | 0 | 3 | 24090 | 24378 | 289 | 0.8368 | 1.8363 | bicycle |
| **4** | 4 | 0 | 4 | 22960 | 23209 | 250 | 0.8188 | 1.8237 | bicycle |
| **...** | ... | ... | ... | ... | ... | ... | ... | ... | ... |
| **6530** | 264 | 1 | 264 | 26011 | 26329 | 319 | 2.0857 | 4.5174 | car |
| **6531** | 265 | 1 | 265 | 26057 | 26397 | 341 | 2.1942 | 4.9713 | car |
| **6532** | 266 | 1 | 266 | 26264 | 26483 | 220 | 2.1259 | 4.8417 | car |
| **6533** | 267 | 1 | 267 | 26295 | 26483 | 189 | 2.1231 | 5.0422 | car |
| **6534** | 268 | 1 | 268 | 10540 | 10901 | 362 | 1.9505 | 4.7414 | car |

6535 rows × 9 columns

**Table two :The roundD dataset identifies pedestrians, bicycles, cars, trucks, and buses.**

## METHODS

Our method consists of a two-step process: First, we record the natural behavior of road users using a drone. To classify erratic behavior, variable scales were used as important safety criteria to distinguish between each driving behavior, which has been used in many studies [25], [26]. But we got the pre-recorded data.

Second, the processing of recordings on a computing cluster, where road users are detected in each video frame and paths are generated from these discoveries.

This is done using cluster analysis by the elbow method to determine the optimal number of clusters, which shows the importance of using the resulting data and for analyzing and processing the recordings, we used K-mean as a clustering algorithm that makes the objects into the same cluster (called cluster), we used Seaborn diagrams to discover and understand the data where the planning functions work Their data frameworks contain a complete data set.

This research paper describes the effect of roundabouts on road users. We customized cars, buses, and tracks, and ignored pedestrians, cyclists, and other situations.

We used samples recorded by a drone, which contains a camera that depicts road users from different angles and at different times of the day on the roundabouts, and we took samples and separated the data of interest to us in this research.

My team and I analyzed the data and extracted the values using the equations shown and got the results also shown in the DVs.

We studied and tracked these results to obtain concepts to use in formulating our research, and we also pooled the data using K-mean as a clustering algorithm.

At the end of our research, we got this research paper, which hopefully will be a useful point in favor of science. The attached Python files explain our way of analyzing the data, and PowerPoint explains the most important details and how we work in this research.

## Volatility Measures

volatility indices can be utilized to identify locations that driving behavior is different compared to driving behavior of same drivers at round intersections.

In this study, volatility functions were applied to Speed volatility, Longitudinal acceleration volatility, Lateral deceleration volatility,vehicular jerk .In order to process and calculate the volatility indices, equations were used in calculating so as to collect kinetic data about the vehicles.

| Volatility Measure | Description | Equation |
|---|---|---|
| $DV_1$ | Standard deviation of speed | $\sqrt{\frac{\sum_{i=1}^{N}(V_i-\bar{V})^2}{N}}$ |
| $DV_2$ | Standard deviation of longitudinal deceleration or acceleration | $\sqrt{\frac{\sum_{i=1}^{N}\left(AD_{long_i}-\overline{AD_{long}}\right)^2}{N}}$ |
| $DV_3$ | Coefficient of variation of speed | $100 \times \frac{\sqrt{\frac{\sum_{i=1}^{N}(V_i-V)^2}{N}}}{V}$ |
| $DV_4$ | Coefficient of variation of longitudinal acceleration | $100 \times \frac{\sqrt{\frac{\sum_{i=1}^{N}\left(A_{long_i}-A_{long}\right)^2}{N}}}{\overline{A_{long}}}$ |
| $DV_5$ | Coefficient of variation of longitudinal deceleration | $100 \times \frac{\sqrt{\frac{\sum_{i=1}^{N}\left(D_{long_i}-D_{long}\right)^2}{N}}}{\overline{D_{long}}}$ |
| $DV_6$ | Mean absolute deviation of speed | $\frac{\sum_{i=1}^{N}\lvert V_i-\bar{V}\rvert}{N}$ |
| $DV_7$ | Mean absolute deviation of longitudinal acceleration | $\frac{\sum_{i=1}^{N}\left\lvert A_{long_i}-\overline{A_{long}}\right\rvert}{N}$ |
| $DV_8$ | Quantile coefficient of variation of normalised speed | $100 \times \frac{Q_{V_3}-Q_{V_1}}{Q_{V_3}+Q_{V_1}}$, where $Q_1$ and $Q_3$ are the sample $25^{th}$ and $75^{th}$ percentiles. |
| $DV_9$ | Quantile coefficient of variation of longitudinal acceleration | $100 \times \frac{Q_{A_{long_3}}-Q_{A_{long_1}}}{Q_{A_{long_3}}+Q_{A_{long_1}}}$ |
| $DV_{10}$ | Quantile coefficient of variation of longitudinal deceleration | $100 \times \frac{Q_{D_{long_3}}-Q_{D_{long_1}}}{Q_{D_{long_3}}+Q_{D_{long_1}}}$ |

**Table 3 : Volatility Measures(Speed,Longitudinal Deceleration,Longitudinal Acceleration,Longitudinal Deceleration or Acceleration).**

**K-means Algorithm**

The global K-means algorithm, which is an incremental approach to clustering algorithm,calculates the centroid and iterates until the optimal centroid is found. The number of combinations found from the method data is denoted by the letter “K” in the K-mean.

In this method, the data points are mapped to the groups in such a way that the sum of the squared distances between the data points and the midpoint is as small as possible (to reduce the error function). It is essential to note that lower diversity within groups leads to more identical data points within the same group.

**The working steps of the K-Means group technique can be described :**

Step 1: First, we need to provide the number of groups, K, that should be generated by this algorithm , k initial "means" (In our search k=3) Created based on (bus, car, truck) within the data domain .

Step 2: Next, choose K data points and assign each one to a group(bus, car, truck) . In short, classify the data based on the number of data points.

Step 3: The cluster centroids will be calculated.

Step 4: Repeat the steps below until we find the perfect central point, which is to map data points to groups that do not differ Just(bus, car, truck).

4.1 The sum of the squared distances between data points and centroids will be calculated first.

4.2 At this point, we need to allocate each data point to the block closest to the other clusters.

4.3 Finally, calculate the centroids of the clusters by averaging all the cluster data points.

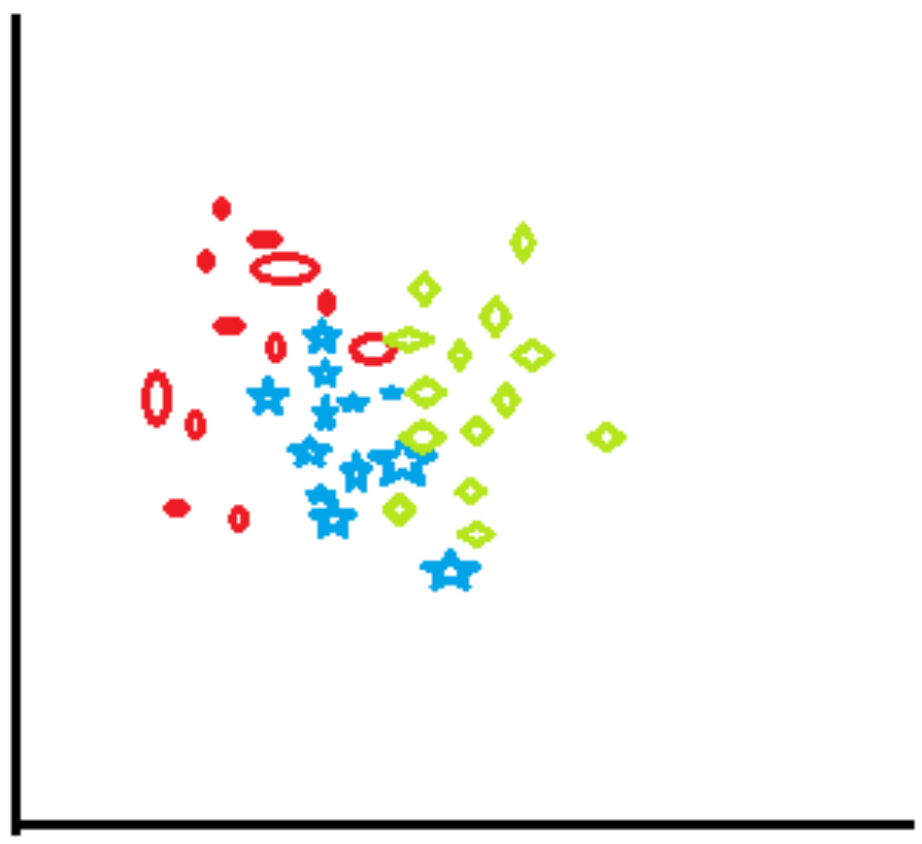

fig 1: before applying k-means clustering

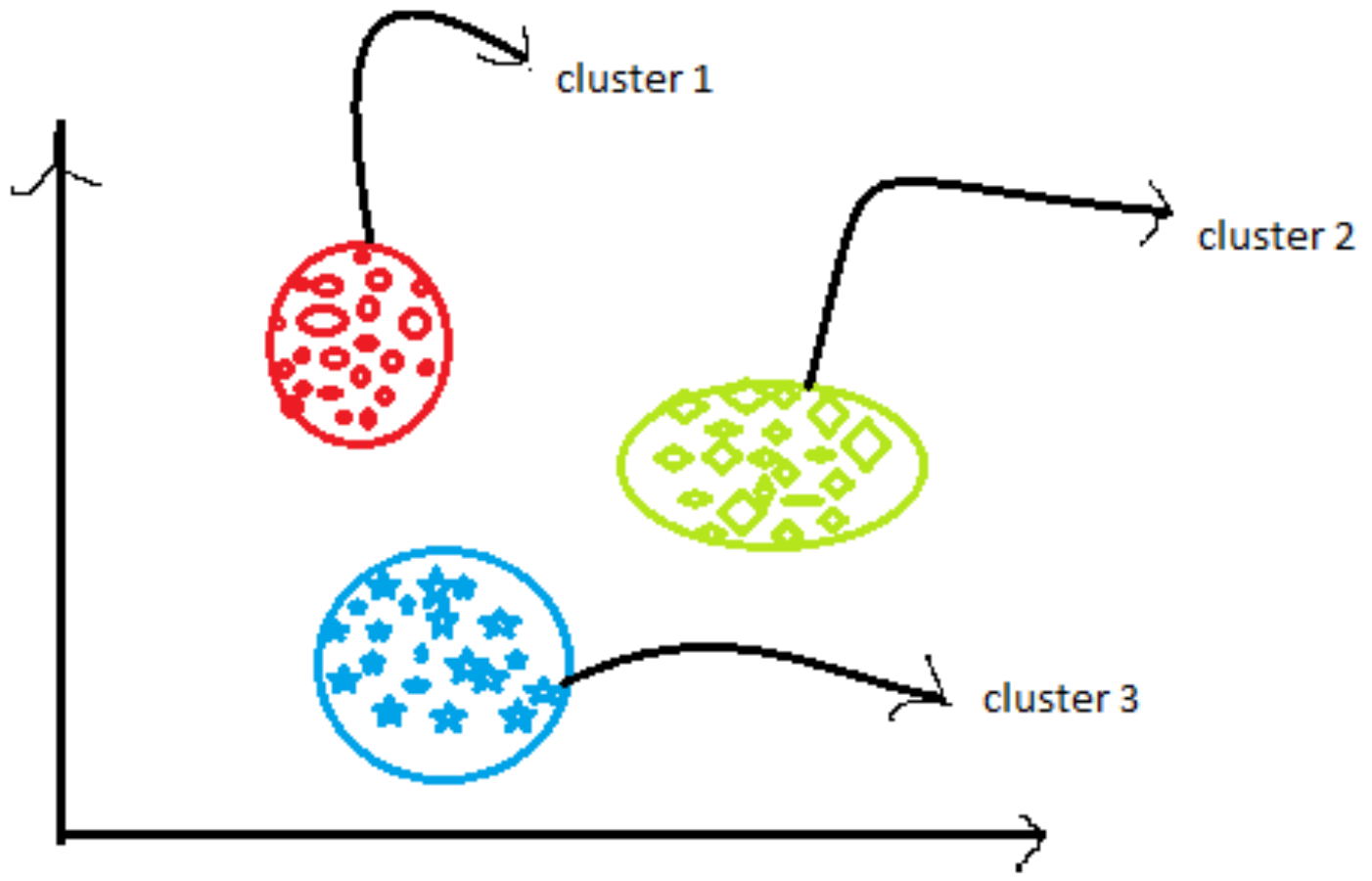

fig 2: After applying K-means clustering

# RESULTS AND DISCUSSION

## Descriptive Analysis

As mentioned earlier also, the main objective of this paper is to show the potential effects of roundabout intersections on pedestrians and roundabout users. In this section, the results of the study are presented in detail in order to verify the practical application of traffic data collection and processing. Data collected via drone flights is used to analyze traffic volume and capacity, examining the difference between each category. In addition, the extracted data is also used to analyze the behavior of drivers. The following subsections provide an in-depth description of the experiment and the entire analytical process:

**Road User Behavio**r: Speed Analysis In this analysis, roundabout traffic behavior is divided into three types: reckless drivers, conservative drivers, and casual drivers. It should be noted that pedestrians and means that use two wheels were excluded, and the speed characteristics of the users were compared. When applying the equations attached below to calculate the speed and deceleration ratio, we noticed a decrease in the speed of the reckless driver at the roundabout, and the conservative driver's commitment to speed was better and closer to the normal imposed speed, and it was closer The two cases is the normal driver, and he had more than one state of commitment and recklessness or was in the average rate. The graphical results of the analysis are shown in the figure below.

### My work limits:

Due to our lack of experience in writing scientific research and analyzing data using Python language, one of the concerns we faced was the validity of our findings, was our representation correct, did we achieve what we wanted, and was able to explain and show what we got in the desired and correct way.

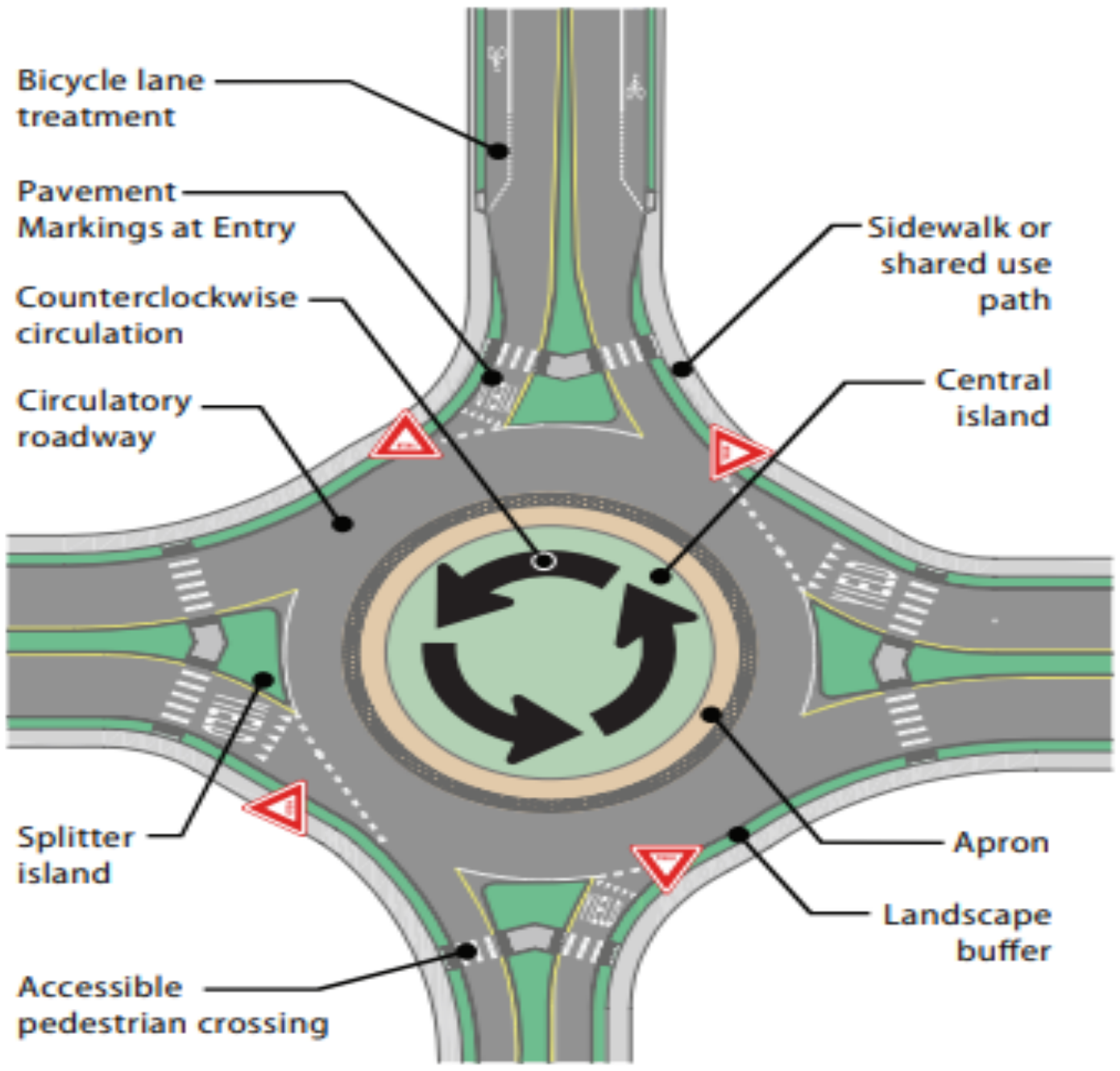

| | count | mean | std | min | 25% | 50% | 75% | max |
|---|---|---|---|---|---|---|---|---|
| **DV1** | 4380.0 | 2.758153 | 1.484332 | 0.225040 | 1.676293 | 2.066563 | 4.022968 | 6.949663 |
| **DV2** | 4380.0 | 0.398081 | 0.165311 | 0.036550 | 0.284555 | 0.377236 | 0.494382 | 1.623159 |
| **DV3** | 4380.0 | 1889.133934 | 1016.659525 | 154.136290 | 1148.138793 | 1415.445387 | 2755.440342 | 4760.012849 |
| **DV4** | 4380.0 | 1736.793134 | 721.235994 | 159.462384 | 1241.490175 | 1645.848721 | 2156.948122 | 7081.705390 |
| **DV5** | 4380.0 | 2417.766760 | 1072.671728 | 68.612753 | 1706.867248 | 2355.544953 | 2994.269233 | 20478.570718 |
| **DV6** | 4380.0 | 2.380422 | 1.435634 | 0.167377 | 1.353844 | 1.685608 | 3.610130 | 6.520590 |
| **DV7** | 4380.0 | 0.334062 | 0.144954 | 0.029960 | 0.238731 | 0.310232 | 0.412435 | 1.145860 |
| **DV8** | 4380.0 | 29.125020 | 17.234620 | 3.426700 | 17.097142 | 24.230756 | 36.799800 | 99.359781 |
| **DV9** | 4380.0 | 44.468578 | 17.882372 | 4.391198 | 30.459851 | 43.597760 | 58.305904 | 93.208673 |
| **DV10** | 4380.0 | -54.275473 | 17.929852 | -98.507790 | -67.628758 | -55.861497 | -38.788679 | -15.106458 |

**Table 4 : Descriptive Statistics for Volatility Measures .**

## Clustering Analysis

**Volatility metrics were extracted and used to classify the drivers**. K-mean methods are applied to group drivers into aggressive, normal and conservative groups. Pooling results indicate that not only does driving style differ between drivers, but the rotors have fewer conflict points. A single roundabout has 50% fewer pedestrian collision points than a signal-controlled intersection or similar stop. The conflict between bicycles and vehicles is also reduced. [27]

According to the results that appeared, the measures of driving volatility have relatively similar values, the mean values of the ten volatility scales are relatively close, especially in the normal and conservative group, and the proportion of normal driving styles was higher.

We categorized the behavior of vehicles (bus, car, truck) and neglected other road users through cluster analysis. We used the results of the equations for measuring the volatility on the behavior of compounds. Finally, through the cluster analysis, the vehicle behaviors were represented by a chart that shows the effect of the rotors on each of (bus, car, truck) independently.

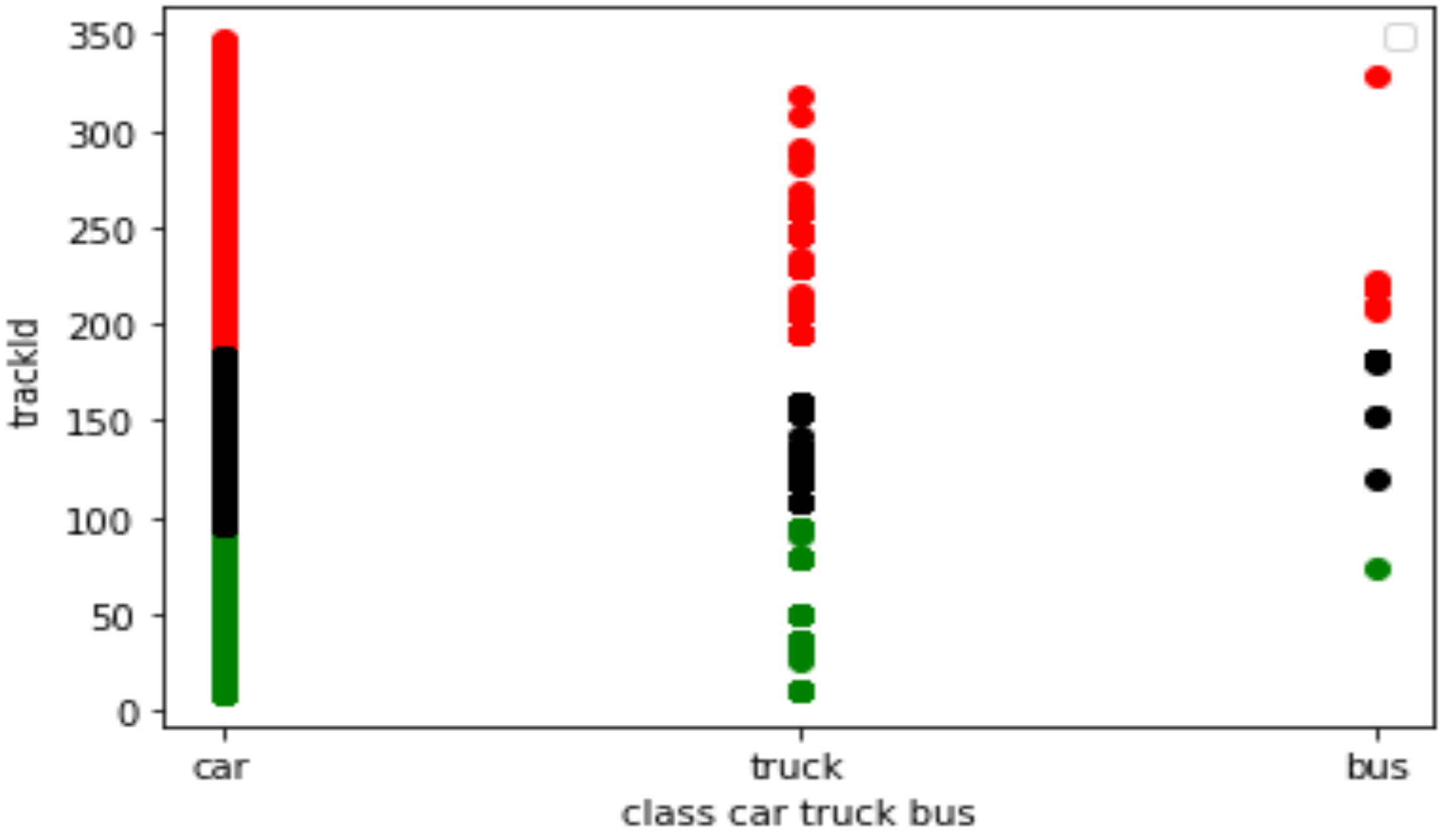

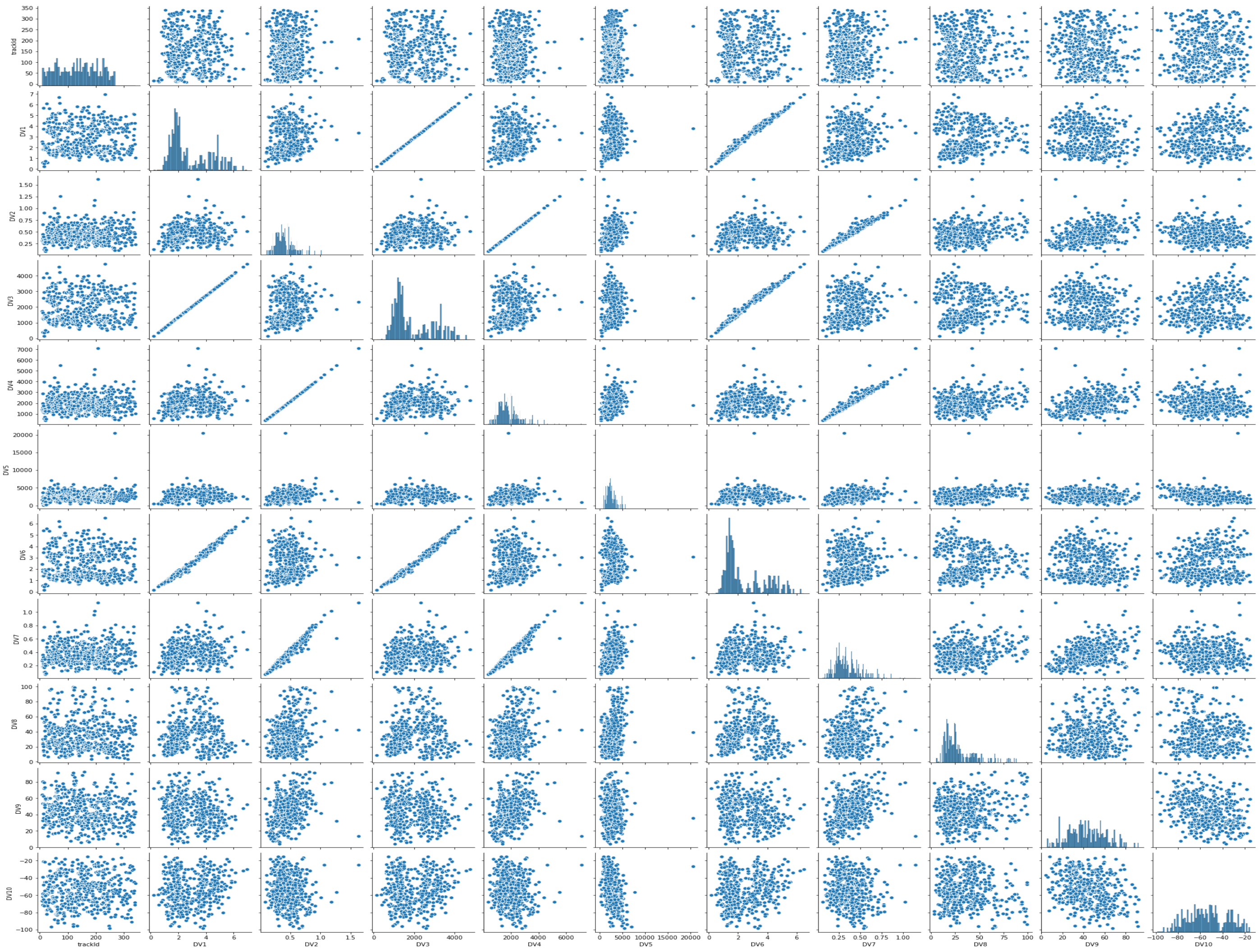

**Figure 1 : Correlation Matrix for All Possible Features.**

# CONCLUSION

The effect of roundabouts on road users, we note at the end of our results, as a result of the positive effect of roundabouts on the reckless driver, who worked to reduce his speed at the beginning of reaching the roundabout, as the roundabouts are important for him. .Regulating traffic and keeping speed within its limits

With the increase in the number of means of transportation and the number of cars on the street, which led to an increase in the number of roundabouts, the importance of the .presence of roundabouts, and the increase in the search for and behavior of roundabouts

Scientists and researchers continue to collect data on rotors and road users to study their .behavior and the interactions between the two ends
If we cannot understand the true impact of roundabouts on road users, our understanding and statistics about the importance of the roundabout, the ways in which a roundabout is developed, how traffic lights are placed in it, or even the way drivers treat it, will decrease. More research and data collection is needed to reach the best point of awareness. in .handling rotors

"